%% file: acl_latex.tex
\pdfoutput=1
\documentclass[11pt]{article}

\usepackage[]{acl}

\usepackage{times}
\usepackage{latexsym}
\usepackage{booktabs} % http://ctan.org/pkg/booktabs
\usepackage{paralist}
\usepackage{xspace}
\usepackage{graphicx}

\usepackage[T1]{fontenc}
\usepackage[utf8]{inputenc}

\usepackage{microtype}

\title{Transformer-based Localization from Embodied Dialog with Large-scale Pre-training}

\author{Meera Hahn\thanks{\enspace Work done in part at Georgia Institute of Technology.} \\
  Google Research \\
  \texttt{meerahahn@google.com} \\\And
  James M. Rehg \\
  Georgia Institute of Technology \\
  \texttt{rehg@gatech.edu} \\}

\begin{document}
\newcommand{\myquote}[1]{\emph{`#1'}}
\newcommand{\xhdr}[1]{\vspace{3pt}\noindent\textbf{#1}}
\newcommand{\Obs}{\textit{Observer}\xspace}
\newcommand{\Loc}{\textit{Locator}\xspace}
\newcommand{\obs}{\textit{observer}\xspace}
\newcommand{\loc}{\textit{locator}\xspace}
\maketitle
\begin{abstract}
We address the challenging task of Localization via Embodied Dialog (LED). Given a dialog from two agents, an Observer navigating through an unknown environment and a Locator who is attempting to identify the Observer's location, the goal is to predict the Observer's final location in a map. We develop a novel LED-Bert architecture and present an effective pretraining strategy. We show that a graph-based scene representation is more effective than the top-down 2D maps used in prior works. Our approach outperforms previous baselines.
\end{abstract}

\input{sections/introduction.tex}
\input{sections/related_work.tex}
\input{sections/approach.tex}

\input{sections/experiments.tex}
\input{sections/conclusion}
\clearpage
\begin{figure*}[hbt!]
\centering
\includegraphics[ width=\textwidth]{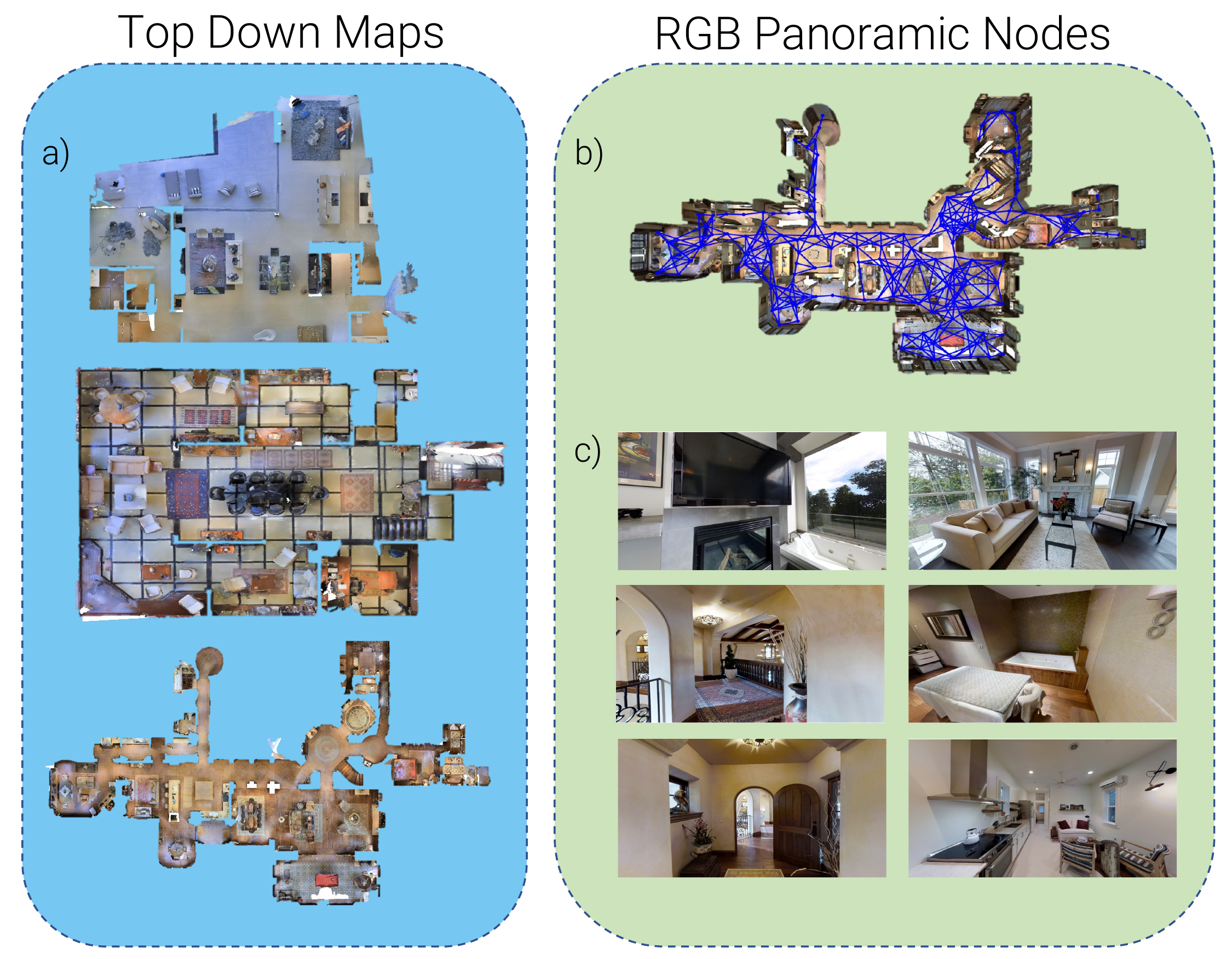}
\caption{Examples of the types of map representations of the Matterport3D~\cite{chang2017matterport3d} indoor environments which can be used for the Localization via Embodied Dialogue task. Part A shows the top down floor maps used in the original LED paper. Part B shows an overlay of the navigation graph of panoramic nodes over the top down map, note the lines represent traversability between nodes and the circles represent the panoramic node location. Part C shows examples of the FPV panoramic nodes in different environments. Note each of these images are mapped to a node in a connectivity graph for the respective environment.}
\label{maps-figure}
\end{figure*}
\bibliography{custom}
\bibliographystyle{acl_natbib}
\input{sections/supplementary}

\end{document}

%% file: sections/introduction.tex
\section{Introduction}
A key goal in AI is to develop embodied agents that can accurately perceive and navigate an environment as well as communicate about their surroundings in natural language. The recently-introduced Where Are You? (WAY) dataset~\citep{hahn2020way} provides a setting for developing such a multi-modal and multi-agent paradigm. This dataset (collected via AMT)  contains episodes of a localization scenario in which two agents communicate via turn-taking natural language dialog: An \emph{Observer} agent moves through an unknown environment, while a \emph{Locator} agent attempts to identify the \textit{Observer's} location in a map. 

The \Obs produces descriptions such as \myquote{I'm in a living room with a gray couch and blue armchairs. Behind me there is a door.} and can respond to instructions and questions provided by the \Loc: \myquote{If you walk straight past the seating area, do you see a bathroom on your right?} Via this dialog (and without access to the \textit{Observer's} view of the scene), the \Loc attempts to identify the \textit{Observer's} location on a map (which is not available to the \Obs). This is a complex task for which a successful localization requires accurate situational grounding and the production of relevant questions and instructions. 

One of the benchmark tasks supported by WAY is `Localization via Embodied Dialog (LED)'. In this task a model takes the dialog and a representation of the map as inputs, and must output a prediction of the final location of the \Obs agent. The model's performance is based on error distance between the predicted location of the \Obs and its true location. LED is a first step towards developing a \Loc agent. One challenge of the task is to identify an effective map representation. The LED baseline from~\citep{hahn2020way} uses 2D images of top down (birds-eye view) floor maps to represent the environment and an (x,y) location for the \Obs.  

\begin{figure}[t]
\centering
\includegraphics[width=1\columnwidth]{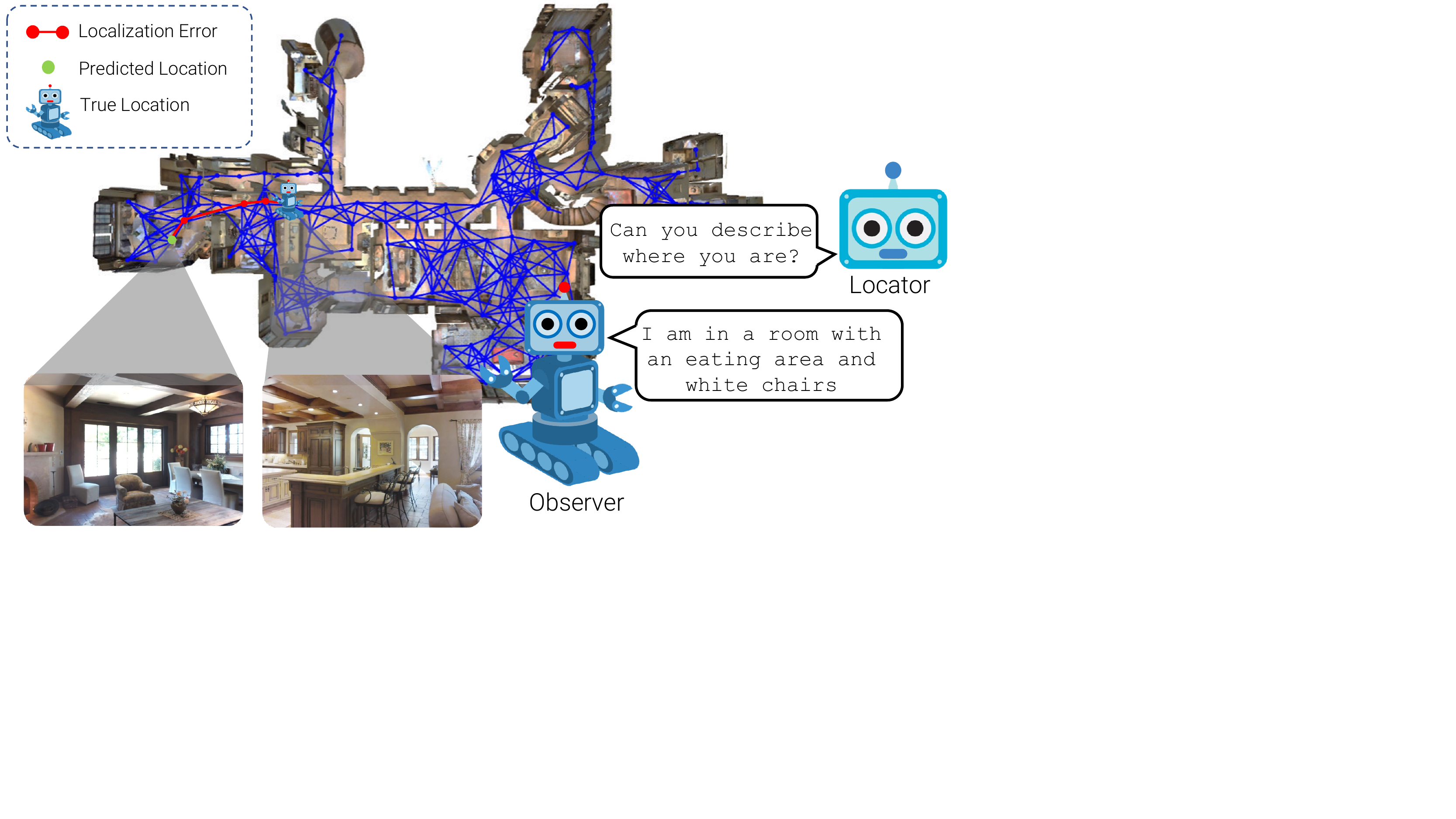}
\caption{WAY Dataset Localization Scenario: The \Loc has a map of the building and is trying to localize the \Obs by asking questions and giving instructions. The \Obs has a first person view and may navigate while responding to the \Loc. The turn-taking dialog ends when the \Loc predicts the \textit{Observer's} position.}
\label{fig:task_teaser}
\end{figure}

This paper provides a new solution to the LED task with two key components. First, we propose to model the environment using the first person view (FPV) panoramic navigation graph from Matterport~\cite{mattersim}, as an alternative to top-down maps. Second, we introduce a novel visiolinguistic transformer model, LED-Bert, which scores the alignment between navigation graph nodes and dialogs. LED-Bert is an adaption of ViLBERT~\citep{lu2019vilbert} for the LED task, and we show that it outperforms all prior baselines. A key challenge is the small size of the WAY dataset (approximately 6K episodes), which makes it challenging to use transformer-based models given their reliance on large-scale training data. We address this challenge by developing a pretraining approach - based on~\cite{majumdar2020improving} - that yields an effective visiolinguistic representation.

\noindent\textbf{Contributions}: To summarize:
\begin{compactenum}
    \item We demonstrate an LED approach using navigation graphs to represent the environment.
    \item We present LED-Bert, a visiolinguistic transformer model which scores alignment between graph nodes and dialogs. We develop an effective pretraining strategy that leverages large-scale disembodied web data and similar embodied datasets to pretrain LED-Bert. 
    \item We show that LED-Bert outperforms all baselines, increasing accuracy at 0m by 8.21 absolute percent on the test split.
\end{compactenum}

%% file: sections/related_work.tex
\section{Related Work}
\xhdr{BERT}
Bidirectional Encoder Representations from Transformers (BERT) is a transformer based encoder used for language modeling. BERT is trained on massive amounts of unlabeled text data, and takes as input sentences of tokenized words and corresponding positional embeddings per tokens. BERT is trained using the masked language modeling and next sentence prediction training objectives. 
In the masked language modeling schema, 15\% of the input tokens are replaced with a [MASK] token. The model is then trained to predict the true value of the input tokens which are masked using the other tokens as context. In the next sentence prediction schema, the model is trained to predicted if the two input sentences follow each other or not. BERT is specifically trained on Wikipedia and BooksCorpus~\citep{zhu2015aligning}. 

\xhdr{ViLBERT}
ViLBERT~\citep{lu2019vilbert} is a multi-modal transformer that extends the BERT architecture~\citep{devlin2018bert} to learn joint visio-linguistic representations. Similar multi-modal transformer models exist~\citep{li2020unicoder, li2019visualbert, su2019vl, tan2019lxmert, zhou2020unified}. ViLBERT is constructed of two transformer encoding streams, one for visual inputs and one for text inputs. Both of these streams use the standard BERT-BASE~\citep{devlin2018bert} backbone. The input tokens for the text stream are text tokens, identical to BERT. The input tokens for the visual stream are a sequence of image regions which are generated by an object detector pretrained on Visual Genome~\cite{krishna2017visual}. The input to ViLBERT is then a sequence of visual and textual tokens which are not concatenated and only enter their respective streams. The two streams then interact using co-attention layers which are implemented by swapping the key and value matrices between the visual and textual encoder streams for certain layers. Co-attention layers are used to attend to one modality via a conditioning on the other modality, allowing for attention over image regions given the corresponding text input and vise versa. 
%Training ViLBERT requires paired image-text data, and ViLBERT specifically uses the Conceptual Captions dataset~\citep{sharma2018conceptual} which is a large dataset of images from the web paired with an alt-text.
%Masked multi-modal modelling and multi-modal alignment tasks are used to train ViLBERT and are used in the pretraining and fine-tuning of LED-Bert. Masked multi-modal modelling works in a similar fashion to masked language modeling in BERT. The multi-modal alignment objective trains ViLBERT to determine if a input image-text pair are well aligned and matched. This alignment objective can be directly extended to matching dialogs and node pairs.

\xhdr{Vision-and-Language Pre-training} 
Prior work has experimented with utilizing dual-stream transformer based models that have been pretrained with self-supervised objectives and transferring them to downstream multi-modal tasks with large success. This has been seen for tasks such as Visual Question Answering~\cite{antol2015vqa}, Commonsense Reasoning~\cite{zellers2019recognition}, Natural Language Visual Reasoning~\cite{suhr2018corpus}, Image-Text Retrieval~\cite{lee2018stacked}, Visual-Dialog~\cite{murahari2020large} and Vision Language Navigation \cite{majumdar2020improving}. Specifically VLN-Bert and VisDial + BERT adapt the ViLBERT architecture and utilize a pretraining scheme which inspired our approach to train LED-Bert.
\vspace{-2mm}

%% file: sections/approach.tex
\section{Approach}
\label{sec:approach}
\vspace{-1mm}

\subsection{Environment Representation}
\label{graph}
A key challenge in the LED task is that environments often have multiple rooms with numerous similar attributes, i.e. multiple bedrooms with the same furniture. Therefore a successful model must be able to visually ground fine-grained attributes. Strong generalizability is also required in order to generalize to unseen test environments. The LED baseline in~\cite{hahn2020way} approaches localization as a language-conditioned pixel-to-pixel prediction task -- producing a probability distribution over positions in a top-down view of the environment, illustrated in Part A, in the Supplementary, Figure 3. This choice is justified by the fact that it mirrors the observations that the human \Loc had access to during data collection, allowing for a straightforward comparison. However, this does not address the question of what representation is optimal for localization.

We propose to use a navigation-graph map representation derived from the panoramic-RGB graphs of the Matterport environments~\cite{chang2017matterport3d}, illustrated in Part B, in the Supplementary, Figure 3. The \Obs agent traverses these same navigation graphs during data collection, which may result in a strong alignment between the dialog and the nodes. Using this approach, the LED task can be framed as a prediction problem over the possible nodes in the navigation graph. At inference time, this can be accomplished by producing an alignment score between each node in the test environment and the test dialog, and then returning the node with the highest score as the predicted \Obs location. 

\vspace{-3mm}
\subsection{Adapting ViLBERT for LED}
To formalize the graph based LED task, we consider a function $f$ that maps a node location $n$ and a dialog $x$ to a compatibility score $f(n,x)$. We model $f(n, x)$ using a visiolinguistic transformer-based model we denote as LED-Bert, shown in Figure ~\ref{led-bert-arch}. The architecture of LED-Bert is structurally similar to ViLBERT and VLN-Bert~\cite{majumdar2020improving}, but with some key differences due to our need to ground dialog and fine-tune on the relatively small WAY dataset. This enables transferring the visual grounding learned during pretraining on disembodied large-scale web data and similar embodied grounding tasks. In the implementation we initialize the majority of LED-Bert using pretrained weights from VLN-Bert. 

The input to the LED-Bert model is a dialog and and a single node from the environment graph map. We represent each panoramic node $I$ as a set of image regions ${r_1,...,r_k}$. We represent an dialog $x$ as a sequence of tokens $w1, . . . , w_L$. Then for a given dialog-node pair the input to LED-Bert is the following sequence:
\setlength{\abovedisplayskip}{3pt}
\setlength{\belowdisplayskip}{3pt}
\begin{equation}
\resizebox{.85\hsize}{!}{$\mbox{\texttt{<IMG>~}} r_1, \dots, r_k \mbox{\texttt{~<CLS>~}} w_1, \dots, w_L \mbox{\texttt{~<SEP>~}}$}
\end{equation}
\noindent where \texttt{IMG}, \texttt{CLS}, and \texttt{SEP} are special tokens. Transformer models are by nature invariant to sequence order and they only model interactions between inputs as a function of their values~\cite{vaswani2017attention}. This leads to the standard practice of adding positional embeddings for each input token to re-introduce order information. For the dialog tokens we simply use an index sequence order encoding. However the panoramic node visual tokens have a more complicated positional encoding, as the panorama is broken up into image regions. The visual positional information is very important for encoding spatial relationships between objects and for scene understanding as a whole. For instance consider the question the \Loc might ask, \myquote{Are you located to the right of the blue couch?} This question will require information about which region of the panorama the couch is located in. We address this by follow the VLN-Bert~\cite{majumdar2020improving} strategy of encoding the spatial location of each image region, ${r_k}$. Each image region is encoded terms of its location in the panorama (top-left and bottom-right corners in normalized coordinates as well as area of the image covered) and its elevation relative to the horizon. Note all angles are encoded as $[\cos(\theta),\sin(\theta)]$. The resulting encoding is an 11-dimensional vector $S$ which is projected into 2048 dimensions using a learned projection $W^S$. 

\vspace{-2mm}
\subsection{Training Procedure for LED-Bert}
\begin{figure*}[ht]
\centering
\includegraphics[ width=\textwidth]{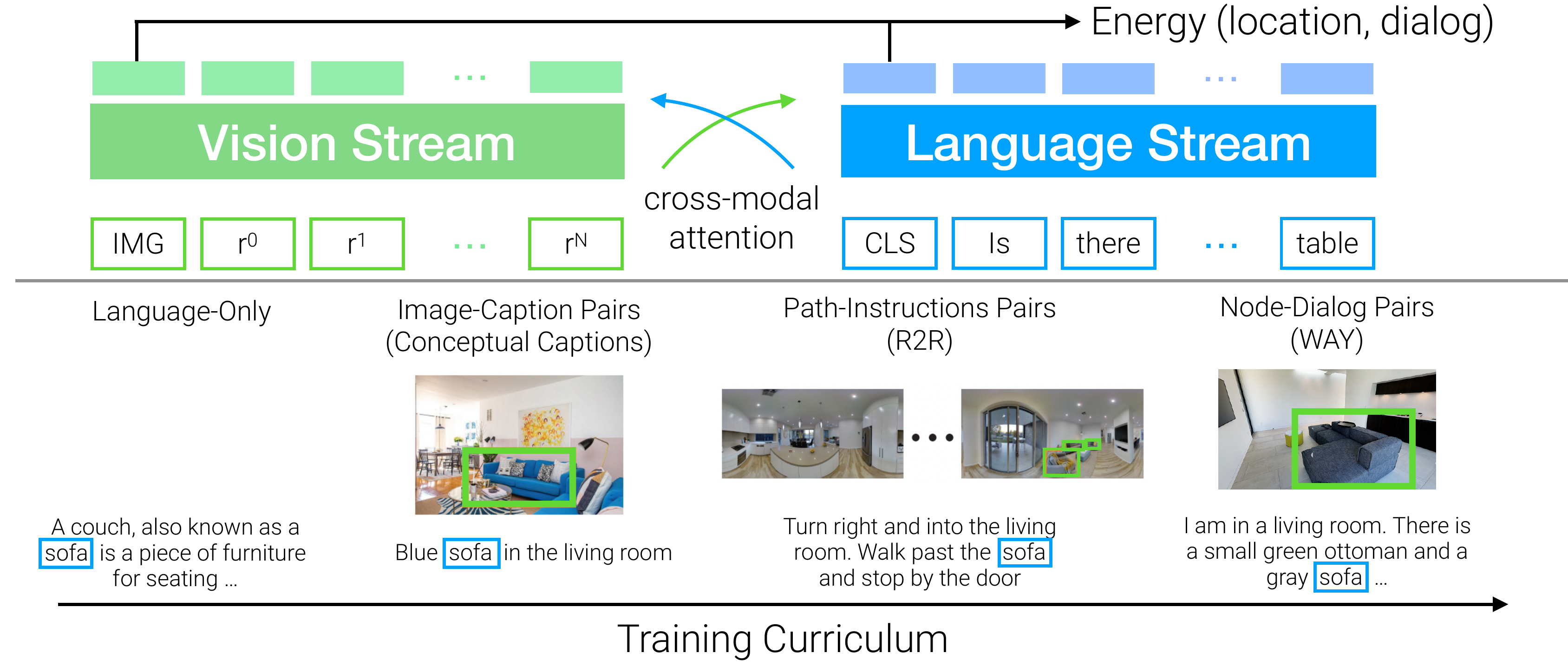}
\caption{We propose the LED-Bert for the LED task. The model is pretrained in 3 stages over different datasets before being fine-tuned over the node-dialog pairs of the WAY dataset~\cite{hahn2020way}. The language stream of the model is first pretrained on English Wikipedia and the BooksCorpus~\cite{zhu2015aligning} datasets. Second, both streams of the model are trained on the Conceptual Captions~\cite{sharma2018conceptual} dataset. Third, both streams are train on the path-instruction pairs of the Room2Room dataset~\cite{anderson2018vision}. Finally we fine-tune the model over the node-dialog pairs of the WAY dataset~\cite{hahn2020way}.}
\label{led-bert-arch}
\end{figure*}

LED-Bert can be trained from scratch using the WAY dataset however due to the small size (~6k episodes) of the WAY dataset and since large-transformer models have been shown to work best on large amounts of data we follow the 4 stage pretraining procedure of prior work~\cite{majumdar2020improving,murahari2020large,lu2019vilbert}. These works do extensive pretraining for multi-modal transformers using large scale web-data. The pipeline for pretraining has 4 stages and is also visualized in Figure ~\ref{led-bert-arch}. 

Stage 1-3 are the same as ~\cite{majumdar2020improving}, and we replace the 4th stage with fine-tuning for node localization over the WAY dataset. To train LED-Bert for localization, we consider the task as a classification task over the possible nodes in the graph, on average there are 117.32 nodes, with the largest environment containing 345 nodes. We run LED-Bert on each node-dialog pair and extract the final representations for each stream, denoted as $h_\texttt{CLS}$ and $h_\texttt{IMG}$, using these we compute a compatibility score by doing element-wise multiplication of the two vectors and passing them through a single linear layer. The scores are normalized via a softmax layer and then supervised using a cross-entropy loss against a one-hot vector with a mass at the ground truth node. 

%% file: sections/experiments.tex
\section{Experiments}

\setlength{\tabcolsep}{5pt}
\begin{table*}[t]
\begin{center}
\caption{Comparison of the LED-Bert model with baselines and human performance on the LED task. We report average localization error (LE) and accuracy at $k$ meters (all $\pm$ standard error).}\vspace{-0.125in}
\footnotesize
\resizebox{\textwidth}{!}{
\label{table:results}
\begin{tabular}{l c ccc c ccc c ccc}
\toprule
                &  & \multicolumn{3}{c}{val-seen}    & & \multicolumn{3}{c}{val-unseen}    & & \multicolumn{3}{c}{test}                                                                                                                                          \\\cmidrule(l{2pt}r{2pt}){3-5} \cmidrule(l{2pt}r{2pt}){7-9}  \cmidrule(l{2pt}r{2pt}){11-13}  
                
Method         & & \multicolumn{1}{c}{\scriptsize LE $\downarrow$}& \multicolumn{1}{c}{\scriptsize Acc@0m $\uparrow$}   & \multicolumn{1}{c}{\scriptsize Acc@5m $\uparrow$}   && \multicolumn{1}{c}{\scriptsize LE $\downarrow$}& \multicolumn{1}{c}{\scriptsize Acc@0m $\uparrow$}   & \multicolumn{1}{c}{\scriptsize Acc@5m $\uparrow$}   && \multicolumn{1}{c}{\scriptsize LE $\downarrow$}& \multicolumn{1}{c}{\scriptsize Acc@0m $\uparrow$}   & \multicolumn{1}{c}{\scriptsize Acc@5m $\uparrow$}   \\ \toprule
Human Locator && 6.00 & 47.87 & 77.38 && 3.20 & 56.13 & 83.42 && 5.89 & 44.92 & 75.00
\\\midrule
Random Node  && 20.8 & 0.33 & 10.82 && 18.61 & 1.9 & 11.05  && 20.93 & 0.92 & 11.00   \\
Center Node && 15.68 & 0.66 & 12.79 && 13.72 & 1.21 & 14.16 && 16.17 & 2.25 & 12.25   \\
LingUNet-Skip && 9.65 & 18.27 & 58.36 && 13.80 & 5.18 & 23.83 && 19.41 & 4.83 & 19.67   \\
Late Fusion    && 12.56 & 17.38 & 47.54 && 12.87 & 7.77 & 34.37 && 15.86 & 8.92 & 32.75    \\
Attention Model     && 9.83 & 18.36 & 56.07 && 10.93 & 10.54 & 41.11 && 14.96 & 6.92 & 34.42 \\
Attention over History Model  && 11.64 & 21.64 & 49.18 && 11.44 & 10.02 & 43.18 && 14.98 & 7.14 & 33.68 \\		
Graph Convolutional Network   && 10.95 & 19.67 & 59.13 && 9.10 & 8.64 & 46.99 && 14.32 & 9.46 & 35.10
\\\midrule		
LED-Bert  && \textbf{9.04} & \textbf{25.57} & \textbf{60.66}  &&  \textbf{8.82} & \textbf{21.07} & \textbf{52.5}   && \textbf{11.12} & \textbf{17.67} & \textbf{51.67}     \\ \bottomrule
\end{tabular}}
\end{center}
\end{table*}

\subsection{Baselines}
We propose a set of strong baseline methods to compare against the LED-Bert architecture. All approaches use the panoramic maps thus ensuring the same prediction space.

\xhdr{Human Performance:} Uses the average performance of AMT \Loc workers from the WAY dataset. We snap the human prediction over the top down map to the nearest node.

\xhdr{Random:} Selects a random node from the test environment as the predicted location for each episode.

\xhdr{Center:} Selects the panoramic node closest to the centroid of the 3D environment point cloud. %In the task set up of Hahn et al., 2020, floor on which the \Obs was located was given as input to the models. This allowed the “center” baseline to be the center pixel of that input floor. In the navigation graph LED task set up the floor is not given and the model must predict over the panoramic nodes across the entire house.

\xhdr{LingUNet-Skip:} Uses the LingUNet-Skip model introduced in the top down floor map task set up of LED~\cite{hahn2020way}. In this set up, the floor on which the \Obs was located was given as input to the models. In the navigation graph LED task set up the floor is not given and the model must predict over the panoramic nodes across the entire house, rather than a single floor. To create a fair comparison between models, we run LingUNet-Skip across all floors in the environment via inputting one floor at a time and then taking the pixel with the highest probability across all floors as the predicted location. We then snap this point to the closest panoramic node and calculate localization error via geodesic distance on the navigation graph. 

\xhdr{Joint Embedding:} This baseline learns a common embedding space between the dialogs and corresponding node locations. Each panoramic node is represented by 36 image patches and image features are extracted for each patch. Visual features are extracted using a ResNet152~\cite{he2015deep} pretrained on Places 365~\cite{zhou2017places}. We experiment with three types of joint embedding architectures - late fusion, dialog based attention, dialog history based attention. All models encode the dialog in the same way and is described below.

\xhdr{Graph Convolutional Network}
Both the joint-embedding baselines and LED-Bert discard edge information. We propose a framework that uses Graph Convolutional Networks (GCN)~\cite{zhang2019graph} to model the LED task using the navigation graph as input which incorporates edge information. 
% GNNs have been used successfully in the embodied settings of visual navigation \cite{ li2019graph, hahn2021no} and the multi-modal settings of visual captioning \cite{pan2020spatio,yang2019auto}. 
In the graph representation input to the model, nodes attributes are visual features and edge attributes contain the pose transformation between connected nodes. 
The goal of the GCN architecture is to model the relational information between the nodes of the graph and the localization dialog in order to produce a probability distribution of localization likelihood over the nodes. 

\xhdr{Dialog Encoding:}
The \Loc and \Obs messages are tokenized using a standard toolkit~\cite{loper2002nltk}. The dialog is represented as a single sequence with identical `start' and `stop' tokens surrounding each message, and then encoded using a single-layer bidirectional LSTM. Word embeddings are initialized using GloVe~\cite{Pennington14glove:global} and fine tuned end-to-end. In the first model called the `late-fusion model', the LSTM has a 2048 dimension hidden state and the node features are down-sampled using self attention to be of size 2048. The visual and dialog features are fused through late fusion passed through a two-layer MLP and softmax and the output is a prediction over the possible nodes in the environment. In the `attention model', the visual and dialog features are fused instead through top-down bottom up attention, the final layers of the model are also an MLP and softmax. In the `attention over history model', there are two separate LSTMs. The former encodes dialog history and the later encodes the current message. Attention via dialog-history is applied over the visual features, then the encoded current message and visual features are fused through late fusion followed by an MLP and softmax.

\subsection{Metrics}
We propose to evaluate the localization error (LE) of our models using geodesic distance instead of euclidean distance as used in \cite{hahn2020way}. Geodesic distance is more meaningful than euclidean distance for determining error across rooms and across floors in multi-story environments. To discern the precision of the models, we report a binary success metric that places a threshold $k$ on the LE. Accuracy (Acc) at 0 meters indicates the correct node was predicted. Accuracy at $k$ meters indicates that the node predicted was within k meters of the true node. 
%This is helpful to understand the precision of the locator model.
\vspace{-3mm}

\subsection{Results}
Table \ref{table:results} shows the performance of our LED-Bert model and relevant baselines on the val-seen, val-unseen, and test splits of the WAY dataset.  

\xhdr{Human and No-learning Baselines.} Humans succeed 44.92\% of the time in test environments at 0 meters; this shows it is a difficult task.

\xhdr{Attention and History increase performance.} Adding bottom-up and top-down attention increases performance, additionally separating the encoders for the current message from the dialog history further increases performance. While it is possible to pretrain the LSTM language encoder, we observe that the common method of using pretrained GloVE~\cite{Pennington14glove:global} embeddings and training the LSTM from scratch is sufficient for learning the language model. 

\xhdr{Graph Networks see slight improvement.} Graph networks see slight increase in performance on the test split. While we believe pretraining the GNN models would boost performance, there is not a straight forward large-scale web-data pretraining schema for the GNN models on this task. 

\xhdr{LED-Bert outperforms all baselines.} LED-Bert significantly outperforms the other cross-modal modeling baselines in terms of both accuracy and localization error -- improving the best baseline, Graph Convolutional Network (GCN), by an absolute 7.54\% (test) to 12.43\% (val-seen and val-unseen). There remains a gap between our model and human performance -- especially on novel environments (-\% vs -\% on test). 

%% file: sections/conclusion.tex
\section{Conclusion}
\label{sec:conclusion}
In summary, we propose a viso-linguistic transformer, LED-Bert, for the LED task and instantiate a new version approach which does localization over the navigation graph. We demonstrate a pre-training schema for LED-Bert which utilizes large scale web-data as well as other multi-modal embodied AI task data to learn the visual grounding required for successful localization's in LED. We show LED-Bert is able to achieve SOTA performance and outperform other learned baselines by a significant margin. 

%% file: sections/supplementary.tex
\section{Supplementary}
\label{sec:appendix}

\subsection{Environment Representation}
The LED baseline in~\cite{hahn2020way} approaches localization as a language-conditioned pixel-to-pixel prediction task -- producing a probability distribution over positions in a top-down view of the environment, illustrated in Part A, Figure ~\ref{maps-figure}.  In this paper we used a navigation-graph map representation derived from the panoramic-RGB graphs of the Matterport environments~\cite{chang2017matterport3d}, illustrated in Part B, Figure ~\ref{maps-figure}.